\documentclass{article}
\usepackage{spconf}
\usepackage{amsmath}
\usepackage{graphicx}
\usepackage{booktabs}
\usepackage{multirow}
\usepackage{xcolor}
\usepackage{algorithm}
\usepackage{algorithmic}
\usepackage{hyperref}
\hypersetup{hidelinks}

\title{DESIGNING TEXTUAL INTERFACES TO ENHANCE LLM REASONING ON AUDIO-VISUAL QUESTION ANSWERING}
\name{Yi Wang$^{1}$, Tianzi Wang$^{2}$, Tong Lei$^{3}$, Dahan Wang$^{3}$, Peter Bell$^{1}$}

\address{
$^{1}$School of Informatics, University of Edinburgh, UK\\
$^{2}$Department of Systems Engineering and Engineering Management,\\
Chinese University of Hong Kong, Hong Kong SAR\\
$^{3}$Key Laboratory of Modern Acoustics, Institute of Acoustics,
Nanjing University, China \\
\texttt{\{Wang.Yi, Peter.Bell\}@ed.ac.uk, twang@se.cuhk.ed.hk}
}
\begin{document}
\ninept
\maketitle
\begingroup
\renewcommand{\thefootnote}{}
\begin{NoHyper}
\footnotetext{This work has been submitted to the IEEE for possible publication.
Copyright may be transferred without notice, after which this version
may no longer be accessible.}
\end{NoHyper}
\endgroup
\begin{abstract}
Automatic prompt optimization (APO) has shown strong effectiveness for text-based LLMs and LLM systems, but its application to audio-visual understanding systems requires suitable textual interfaces through which multimodal reasoning can be exposed to text-based optimizers.
We investigate how the design of textual interfaces affects prompt optimization for multimodal reasoning, on the audio-visual question answering task. We compare three interface designs, including a direct inference system consisting of a single omni-modal large language model (OmniLLM), a captioner-reasoner system, and an agentic architecture in which a primary agent orchestrates subagent OmniLLMs. Experiments with Qwen3-Omni and Gemini show that direct prompt optimization provides model-dependent gains; the captioner-reasoner system and the agentic system expose more interpretable optimization interfaces but are harder to optimize, where error analysis of cases affected by optimization is conducted.
Our study highlights textual interface design as an important factor in making prompt optimization effective for omni-modal reasoning systems. Our implementation details are specified in GitHub, with code also being released there \footnote{\url{https://github.com/yiwang454/interface_AVQA}}. 
\end{abstract}

\section{Introduction}
\label{sec:intro}
\begin{figure}[t]
    \centering
    \includegraphics[width=0.95\columnwidth]{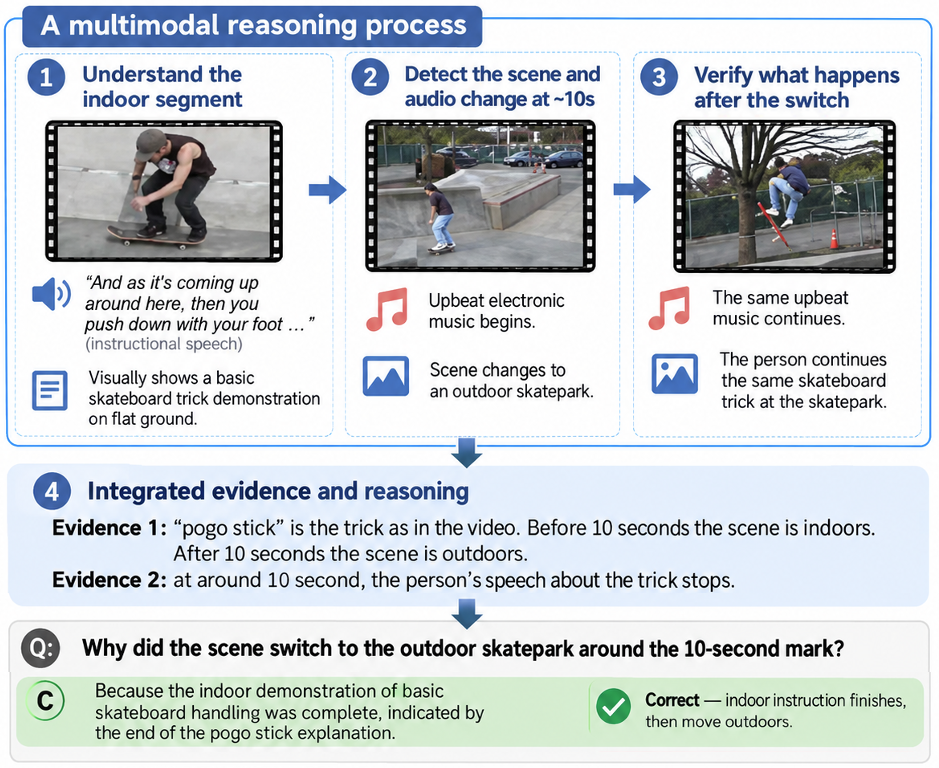}
    \vspace{-0.5em}
    \caption{Demonstration of LLM reasoning on AVQA.}
    \label{fig:demo}
\vspace{-0.2cm}
\end{figure}

\vspace{-0.1cm}
Multimodal reasoning requires integrating evidence from multiple modalities to support a decision. In audio-visual question answering (AVQA) task, for example, this may involve identifying relevant evidence from both audio and video, and reasoning over the evidence to answer a question, as illustrated in Fig. \ref{fig:demo}.
Prior work has improved multimodal reasoning by enhancing chain-of-thought in MLLMs \cite{zhang2023multimodalCOT, gao2025interleaved},
while another approach is to verbalize multimodal evidence into text for subsequent language-based reasoning, where rich and faithful textual descriptions, combined with additional reasoning by text LLMs, have been shown to improve downstream performance \cite{zhang2024simple, gkountouras2025clarification, kumar2026tac, ma2026omni}.
These findings motivate viewing text as an explicit interface through which multimodal reasoning can be structured and improved.
 


Recent omni-modal large language models (OmniLLMs), which jointly process audio, video, and text, have achieved strong performance on audio-visual multimodal reasoning tasks \cite{tang2025videosalmonn2,xu2025qwen3omni,yang2025humanomniv2, comanici2025gemini,team2026qwen3}.
However, their outputs can be sensitive to how the task is instructed. Different prompt instructions for the same audio-visual input may elicit different perceptual details or organize the resulting evidence differently \cite{romero2024question, liu2026prompt}, which may make manual prompt design brittle. 
Automatic prompt optimization (APO) addresses this problem by using task feedback to iteratively search for refined instructions. Given proper optimization signals and evidence, they usually propose prompt that performs better on downstream tasks than heuristic designs \cite{khattab2024dspy, opsahl2024optimizing, agrawal2025gepa}.

\begin{figure}[t]
    \centering
    \includegraphics[width=0.9\columnwidth]{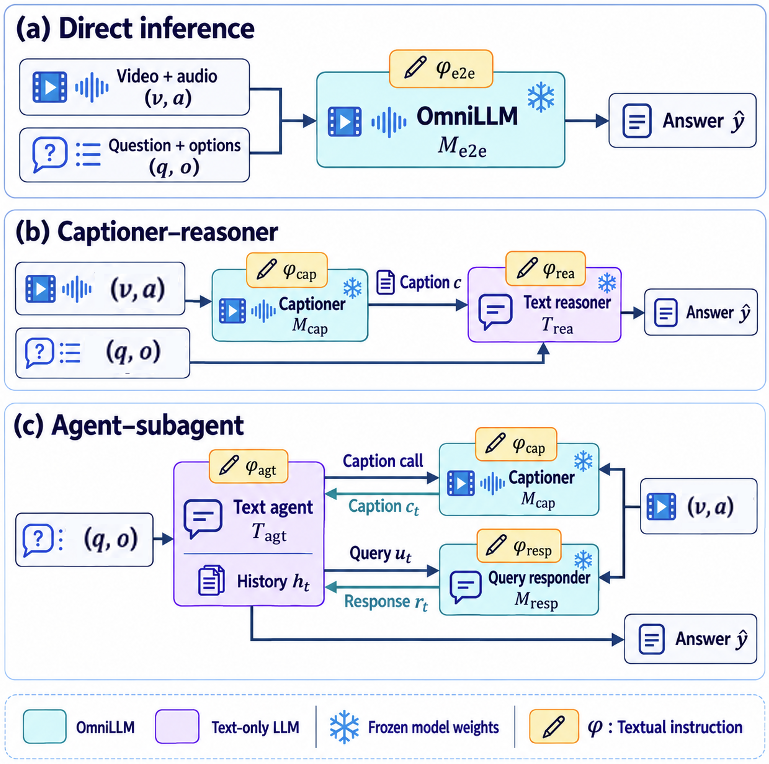}
    \vspace{-0.5em}
    \vspace{-4pt}
    \caption{\textbf{Overview of the three systems we investigated and compared for instruction optimization}.}
    \label{fig:systems}
\vspace{-0.2cm}
\end{figure}

APO methods optimize textual prompts based on text feedback \cite{khattab2024dspy, opsahl2024optimizing, agrawal2025gepa}. Recent multimodal APO methods \cite{liu2026prompt, choi2026multimodal} may additionally incorporate multimodal feedback, but the optimized prompts themselves are still represented in text. Therefore, applying APO to a multimodal reasoning system requires the system to expose a textual optimization interface, specifying which textual instructions the optimizer can modify and what feedback or intermediate information is available to guide prompt updates.

The designs of vision-language interfaces for prompt optimization have been studied in previous works \cite{choi2026multimodal, liu2026prompt}, but less on omni-modal reasoning over audio-visual information.
To this end, we raise the research question: 
does exposing intermediate multimodal reasoning through explicit textual interfaces make APO more effective than optimizing an end-to-end OmniLLM instruction directly?
Thus we explore two types of audio-visual reasoning systems consisting of multi-LLMs, providing different explicit textual optimization interfaces, 1) a \textbf{captioner-reasoner system} that externalizes multimodal evidence into a textual caption before a separate language model produce an answer to the downstream task; 
2) an \textbf{agentic system} where a text-based LLM agent raises queries to OmniLLM subagents to collect multimodal evidence for downstream task. And we compare them with the \textbf{direct inference} system, where a single OmniLLM directly consumes multimodal input to conduct the downstream task.

We study these questions through AVQA benchmarks, where a model must answer questions that depend jointly on the auditory and visual streams of a video \cite{zhou2025daily,yang2025audio,hong2025worldsense}. AVQA provides a natural testbed because it requires both retrieving sparsely distributed multimodal evidence and reasoning over the retrieved evidence. 
Both systems we study are established paradigms for video QA (VQA) or AVQA. Previous works \cite{zhang2024simple, romero2024question, galanopoulos2025llm} conducted caption-based VQA/AVQA focusing on properly aggregating captions from frames, inducing question-conditioned descriptions, etc. Agentic video systems instead acquire question-relevant evidence iteratively through OmniLLM observation and tool use \cite{wang2024videoagent,zhang2026deep,wang2025active, tao2025omniagent}, achieving strong performances. We therefore study captioner-reasoner and agentic systems as representative interfaces for \emph{static evidence externalization} and \emph{dynamic evidence acquisition}, respectively.

While prior work primarily designs such systems to improve task performance, we instead study how effectively their intermediate textual interfaces interact with APO, compared with direct OmniLLM inference. To our knowledge, this is the first systematic study of textual interface design for APO in audio-visual reasoning systems.
Through controlled experiments and error analyses of reasoning trajectories and action traces, we identify distinct limitations in captioner-reasoner and agentic systems, providing insights into further investigation on design and optimization of textual interfaces for audio-visual reasoning.

\vspace{-0.2cm}
\section{Method}
\label{sec:method}
\vspace{-0.2cm}

In AVQA, the audio-visual reasoning system aims to predict the correct answer $y$, given a video $v$ with audio $a$, a question $q$, and answer options $o$. This section introduces three system architectures that differ in how audio-visual evidence is exposed through textual interfaces and, consequently, where
textual instructions can be optimized. All LLM parameters are kept frozen. Three systems consist of OmniLLMs and optionally text-only LLMs. In following sections, $\mathcal{M}$ denotes an OmniLLM and $\mathcal{T}$ denotes a text-only LLM, with subscripts indicating their functional roles. $\phi$ denotes each optimizable prompt, belonging to its corresponding LLM component constructing different systems. 

\vspace{-0.1cm}
\subsection{Direct OmniLLM Inference}
\label{ssec:method-direct}
\vspace{-0.1cm}

We first consider the simplest setting, where an end-to-end (e2e) OmniLLM directly performs AVQA. The model jointly receives the video $v$, audio $a$, question $q$, and answer options $o$, and makes prediction \(\hat y = \mathcal{M}_{\mathrm{e2e}}(v,a,q,o;\phi_{\mathrm{e2e}})\).
$\mathcal{M}_{\mathrm{e2e}}$ denotes the OmniLLM used for end-to-end inference and $\phi_{\mathrm{e2e}}$ is its textual task instruction, informing the model that it should conduct the AVQA task and optionally the procedures and rules to follow.
This system exposes the direct-inference textual interface $\phi_{\mathrm{e2e}}$.

\vspace{-0.1cm}
\subsection{Captioner-reasoner System}
\label{ssec:method-captioner-reasoner}
\vspace{-0.1cm}

We next consider a two-stage captioner-reasoner system.
The captioner produces 
\(c = \mathcal{M}_{\mathrm{cap}}(v,a;\phi_{\mathrm{cap}})\), without knowing the question and options. $\mathcal{M}_{\mathrm{cap}}$ denotes the OmniLLM captioner, and $\phi_{\mathrm{cap}}$ is its textual instruction. A text-only reasoner then predicts 
\(\hat{y} = \mathcal{T}_{\mathrm{rea}}(q,o,c;\phi_{\mathrm{rea}})\), where $\mathcal{T}_{\mathrm{rea}}$ denotes the reasoning LLM and $\phi_{\mathrm{rea}}$ its instruction.
This system exposes two optimizable instructions: the captioner instruction $\phi_{\mathrm{cap}}$ and the reasoner instruction
$\phi_{\mathrm{rea}}$. 

\vspace{-0.1cm}
\subsection{Agent-Subagent System}
\label{ssec:method-agent}
\vspace{-0.1cm}

We further consider an agent-subagent system where the text-based LLM agent interactively acquires audio-visual evidence from OmniLLM subagents, and then makes the answer decision. 
The primary agent cannot inspect the raw video or directly answer the question without calling a subagent. However, it provides longer context window length not consumed by the large amount of multimodal input, allowing itself to memorize information collected from multiple interaction turns with subagents and reason from such memory.

We adopt one subagent for captioning and another for responding to multimodal evidence-collection queries issued by the text agent. The captioner summarizes globally salient events and provides temporal information about these events, while details omitted from the caption can be further collected through targeted queries, conditioned on the current evidence in captions. These two types of tools are widely adopted in agentic audio and video understanding systems and have demonstrated their effectiveness \cite{tao2025omniagent} .

At each interaction step \(t\), the text agent \(\mathcal{T}_{\mathrm{agt}}\) takes an action based on \(q\), \(o\), and interaction history \(h_t\).
The workflow always begins with a captioning call to obtain a description of the audio-visual content.
At each subsequent step, \(\mathcal{T}_{\mathrm{agt}}\) decides its next action:
\(z_t = \mathcal{T}_{\mathrm{agt}}(q,o,h_t;\phi_{\mathrm{agt}}).\)
The action \(z_t\) can be requesting another caption, raising a follow-up query \(u_t\) to a query-response component to collect information useful for answering the original question, or producing the final answer once adequate evidence has been collected.
Both the captioning component and the query-response component are OmniLLMs, just playing different roles.
The captioner component is defined as
\(c_t = \mathcal{M}_{\mathrm{cap}}(v,a;\phi_{\mathrm{cap}})\),
and the follow-up query responder component as
\(r_t = \mathcal{M}_{\mathrm{resp}}(v,a,u_t;\phi_{\mathrm{resp}})\).

This system exposes multiple textual prompts for optimization, most notably $\phi_{\mathrm{agt}}$. The perception-side prompts $\phi_{\mathrm{cap}}$ and $\phi_{\mathrm{resp}}$ are also optimizable, and can be adaptively raised by the text agent.


\vspace{-0.1cm}
\subsection{Automatic Prompt Optimization}
\label{ssec:method-APO}
\vspace{-0.1cm}

During optimization, the system is executed on training examples, whose ground truths are available, using the current instruction $\phi$. The optimizer $\mathcal{F}$, which is a text-based LLM, observes optimization evidence $e$ together with the predicted answer $\hat{y}$ and ground-truth answer $y$, and proposes a revised instruction,
\begin{equation}
    \phi' = \mathcal{F}_{\mathrm{opt}}(\phi,e, y, \hat{y}).
\end{equation}
The evidence may include the model response, intermediate reasoning
trajectory, or, for the agent-subagent system, the interaction trace
containing agent actions and subagent observations.

The optimizer iteratively searches over candidate instructions using the training examples. We use QA accuracy as the task-level optimization objective for all APO algorithms. Candidate instructions produced during the search are evaluated on the validation set,
and the instruction with the highest validation accuracy within the specified optimization budget is selected as the final optimized instruction, to be evaluated on the held-out test set.
\vspace{-0.1cm}
\section{Experiments}
\label{sec:experiments}
\vspace{-0.1cm}
We evaluate APO for AVQA on Daily-Omni \cite{zhou2025daily}, comparing direct single-MLLM inference, captioner-reasoner pipelines, and agentic system, particularly their compatibilities with an APO algorithm GEPA \cite{agrawal2025gepa}, with brief reference experiment on APO algorithm COPRO \cite{opsahl2024optimizing}.

\vspace{-0.1cm}
\subsection{Evaluation Dataset}
\label{ssec:exp_data}

We evaluate on Daily-Omni \cite{zhou2025daily}, which contains 1,197 audio-visual multiple-choice QA examples spanning six task categories. Since APO requires examples for prompt search and validation-based candidate selection, we stratify the benchmark into 125 training, 125 validation, and 947 held-out test examples,
preserving the overall task-category distribution. 
The training set is used during prompt optimization, the validation set for selecting candidate instructions, and the held-out test set as the primary measure of optimization effectiveness. We additionally report full-benchmark accuracy in case for comparison.

\vspace{-0.1cm}
\subsection{Models and Systems}
\label{ssec:exp_models}

For direct single-MLLM inference, we evaluate Qwen3-Omni-30B-A3B in both its Instruct\footnote{Model from: \url{https://huggingface.co/Qwen/Qwen3-Omni-30B-A3B-Instruct}} and Thinking\footnote{Model from: \url{https://huggingface.co/Qwen/Qwen3-Omni-30B-A3B-Thinking}} variants, as well as Gemini-2.5-Flash\footnote{Model information: \url{https://ai.google.dev/gemini-api/docs/models/gemini}}. Qwen3-Omni is served locally, while Gemini-2.5-Flash is accessed via the Google API. 
For the captioner-reasoner system, Gemini-2.5-Flash generates an audio-visual caption and GPT-4.1\footnote{Model card: \url{https://openai.com/index/gpt-4-1/}} performs text-only reasoning. The agent-subagent system similarly uses GPT-4.1 as the primary text agent and Gemini-2.5-Flash for audio-visual perception.
We initially tested Qwen3-Omni variants as captioners, but did not carry them forward to these two systems because they exhibited brittle instruction-following behavior in applying APO for caption instructions, presenting repetitive and meaningless captions.

\vspace{-0.1cm}
\subsection{APO Algorithms}
\label{ssec:exp_optimization}

We evaluate two APO algorithms: COPRO \cite{opsahl2024optimizing}
as a simpler APO for reference method and GEPA \cite{agrawal2025gepa} as the
primary reflective optimizer. GPT-4.1 serves as the optimizer LLM
for both methods. To make their search effort comparable, we use
the same optimization budget of 2,500 trials for COPRO and GEPA.
Candidate instructions are evaluated on the validation set, and the
best validation-performing instruction across the corresponding
search runs is selected for held-out evaluation. Full optimization
and inference configurations are provided in our GitHub repository \footnote{\url{https://github.com/yiwang454/interface_AVQA}}.

\begin{table}[t]
\centering
\caption{Direct QA prompt optimization on Qwen3-Omni.
$\Delta$ Test is relative to the corresponding unoptimized model.}
\label{tab:qwen_direct_opt}
\vspace{2pt}
\small
\setlength{\tabcolsep}{3.2pt}
\resizebox{\columnwidth}{!}{
\begin{tabular}{llcccc}
\toprule
\multicolumn{1}{c}{\multirow{2}{*}{Model}}                                     & \multicolumn{1}{c}{\multirow{2}{*}{Optimizer}} & \multicolumn{4}{c}{Accuracy (\%)}            \\
\multicolumn{1}{c}{}                                                           & \multicolumn{1}{c}{}                           & Val (125) & Overall & Test (947) & $\Delta$ Test \\
\midrule
\multirow{3}{*}{\begin{tabular}[c]{@{}l@{}}Qwen3-Omni\\ Instruct\end{tabular}}
 & None  & 72.80 & 72.18 & \textbf{71.81} & - \\
 & COPRO & 71.20 & 71.93 & 71.59 & -0.22 \\
 & GEPA  & 75.20 & 72.01 & 71.49 & -0.32 \\
\midrule
\multirow{3}{*}{\begin{tabular}[c]{@{}l@{}}Qwen3-Omni\\ Think\end{tabular}}
 & None  & 69.60 & 69.76 & 69.06 & - \\
 & COPRO & 71.20 & 69.26 & 68.85 & -0.21 \\
 & GEPA  & 72.80 & 69.92
         & \textbf{70.01} & +0.95 \\
\bottomrule
\end{tabular}
}
\parbox{\columnwidth}{\footnotesize
val (125) denotes the validation set containing 125 samples. Test (947) denotes the test set containing 947 samples. Same for all other tables. 
}
\vspace{2.5mm}
\end{table}

\begin{table}[t]
\centering
\caption{Direct QA prompt optimization on Gemini-2.5-Flash.}
\label{tab:gemini_direct_opt}
\vspace{2pt}
\small
\setlength{\tabcolsep}{3.5pt}
\resizebox{\columnwidth}{!}
{
\begin{tabular}{llcccc}
\toprule
\multicolumn{1}{c}{\multirow{2}{*}{Model}}
& \multicolumn{1}{c}{\multirow{2}{*}{Optimizer}}
& \multicolumn{4}{c}{Accuracy (\%)} \\
&
& Val (125) & Overall & Test (947) & $\Delta$ Test \\
\midrule

\multirow{3}{*}{
\begin{tabular}[c]{@{}l@{}}Gemini\\ 2.5-Flash\end{tabular}
}
& \multirow{1}{*}{None}
& 77.60
& 77.03
& 75.71
& - \\

& COPRO
& 72.80
& 78.11
& 78.25
& 2.54 \\

& GEPA
& \textbf{79.20}
& \textbf{79.62}
& \textbf{79.83}
& 4.12 \\


\bottomrule
\end{tabular}
}
\vspace{2.5mm}
\end{table}


\begin{table}[t]
\centering
\caption{Effect of the captioner-reasoner interface and prompt optimization.
The direct Gemini inference result is included as a reference. Gemini stands for Gemini-2.5-Flash model.}
\vspace{2pt}
\label{tab:caption_reasoner_interface}
\small
\setlength{\tabcolsep}{3pt}
\renewcommand{\arraystretch}{1.08}

\resizebox{\columnwidth}{!}{
\begin{tabular}{llllccc}
\toprule
\multicolumn{1}{c}{\multirow{2}{*}{System}} &
\multicolumn{1}{c}{\multirow{2}{*}{Captioner}} &
\multicolumn{1}{c}{\multirow{2}{*}{Reasoner}} &
\multicolumn{1}{c}{\multirow{2}{*}{\shortstack{Optimizable\\Component}}} &
\multicolumn{3}{c}{Accuracy (\%)} \\
\cmidrule(lr){5-7}
& & & &
\multicolumn{1}{c}{Val 125} &
\multicolumn{1}{c}{Overall} &
\multicolumn{1}{c}{Test 947} \\
\midrule

\multicolumn{7}{l}{\textit{Direct Inference system as baseline}} \\
Gemini
& \multicolumn{1}{c}{--}
& \multicolumn{1}{c}{--}
& \multicolumn{1}{c}{--}
& \textbf{77.60}
& \textbf{77.03}
& \textbf{75.71} \\

\midrule
\multicolumn{7}{l}{\textit{Same Caption, Reasoner Control}} \\
\multirow{2}{*}{\shortstack{Captioner-\\Reasoner}}
& \multirow{2}{*}{Gemini}
& Gemini
& \multirow{2}{*}{None}
& 68.00
& 69.17
& 69.27 \\
&
&
GPT-4.1
&
& 68.00
& 69.50
& 70.01 \\

\midrule
\multicolumn{7}{l}{\textit{Captioner-Reasoner Optimization}} \\
\multirow{2}{*}{\shortstack{Captioner-\\Reasoner}}
& \multirow{2}{*}{Gemini}
& \multirow{2}{*}{GPT-4.1}
& \shortstack{GEPA\\on Caption}
& 79.20
& 68.50
& 67.90 \\
&
&
&
\shortstack{GEPA\\on Reasoner}
& 71.20
& 70.01
& 69.90 \\

\bottomrule
\end{tabular}
}
\end{table}

\begin{table}[t]
\centering
\caption{Effect of the agentic system and prompt optimization on the text agent workflow instruction.}
\label{tab:react}
\vspace{2pt}
\small
\setlength{\tabcolsep}{5pt}
\renewcommand{\arraystretch}{1.08}
\scalebox{0.9}{
\begin{tabular}{llccc}
\toprule
\multicolumn{1}{c}{\multirow{2}{*}{System}} &
\multicolumn{1}{c}{\multirow{2}{*}{Optimization}} &
\multicolumn{3}{c}{Accuracy (\%)} \\
\cmidrule(lr){3-5}
& & 
\multicolumn{1}{c}{Val 125} &
\multicolumn{1}{c}{Overall} &
\multicolumn{1}{c}{Test 947} \\
\midrule

Gemini
& None
& \textbf{77.60}
& \textbf{77.03}
& 75.71 \\
\midrule
\multirow{2}{*}{ReAct}
& None
& 74.40
& 76.11
& \textbf{76.03} \\

& GEPA
& 76.80
& 75.77
& \textbf{76.03} \\

\bottomrule
\end{tabular}}
\end{table}

\vspace{-0.2cm}
\section{Results and Analysis}
\label{sec:results}

\subsection{APO on the Direct-Inference Interface}
\label{ssec:results_direct}
\vspace{-0.1cm}

We first evaluate whether automatic prompt optimization (APO) can improve the direct QA interface of an OmniLLM. 
Tables~\ref{tab:qwen_direct_opt} and~\ref{tab:gemini_direct_opt} report the results on Qwen3-Omni and Gemini-2.5-Flash, respectively.

For Qwen3-Omni, neither COPRO nor GEPA consistently improves direct QA performance. On Qwen3-Omni-Instruct, both optimizers slightly reduce test accuracy relative to the unoptimized prompt.
Qwen3-Omni-Think exhibits a similar pattern: results after executing COPRO and GEPA is comparable with baseline accuracy on the test set, suggesting that exposing a more explicit chain-of-thought reasoning process does not make APO more effective on the direct interface.

In contrast, Gemini-2.5-Flash (abbreviated to Gemini) benefits clearly from GEPA. The test accuracy improves from $75.71\%$ to $79.83\%$ (by 4.12\%), with overall accuracy also showing an increase. COPRO also brings about an accuracy gain of $2.54\%$.
The performance gain obtained by APO on Gemini may be due to the stronger instruction-following capacity of the model, compared to with the Qwen3-Omni models. We therefore primarily use Gemini as the OmniLLM, and GEPA as the APO due to its more stable performance gain across the validation and test splits, in the following case studies on the other two reasoning systems.

\vspace{-0.2cm}
\subsection{APO on the Captioner-Reasoner Interface}
\label{ssec:results_caption_reasoner}
\vspace{-0.1cm}
The results of experimenting APO on the captioner-reasoner system are shown in Table~\ref{tab:caption_reasoner_interface}.
The captioner-reasoner decomposition itself incurs a substantial performance drop compared with direct Gemini inference. Using the same Gemini-generated captions, both Gemini and GPT-4.1 reasoners obtain around 5\% accuracy
less than the direct inference. The discrepancy between the reasoner and captioner models did not lead to further accuracy drops. 
The drop is possibly because caption forms an information bottleneck between perception and reasoning: the captioner must compress the audio-visual input to text descriptions without knowing which details are key to the question. The downstream reasoner has to make decision on ambiguous observations. 

Moreover, providing additional textual interfaces, from only $\phi_{e2e}$ to $\phi_{cap}$ and $\phi_{rea}$, does not make this architecture easier to optimize. Applying APO to either the caption instruction or the reasoner instruction fails to improve accuracy on the held-out test split. In both cases, validation accuracy increases more strongly than test accuracy. For caption optimization, the validation accuracy rises drastically, while test accuracy decreases from $2.01\%$ This large discrepancy suggests that the captioning interface is especially brittle to prompt changes, making it an unreliable target for APO in this setting. Optimizing the reasoner is stable, but does not provide improvement.

\begin{table}[t]
\centering
\caption{Selected excerpts from GEPA-optimized instructions for direct
Gemini inference and the ReAct agent. }
\label{tab:optimized_prompt_examples}
\small
\setlength{\tabcolsep}{4pt}
\renewcommand{\arraystretch}{1.08}
\resizebox{\columnwidth}{!}{
\begin{tabular}{p{0.16\columnwidth} p{0.55\columnwidth} p{0.22\columnwidth}}
\toprule
\textbf{Interface} &
\textbf{Selected optimized-instruction excerpt} &
\textbf{Main scope} \\
\midrule

Direct
&
``Using only what is directly seen or heard in the video
\ldots identify the single answer choice \ldots best supported by
the video evidence at the relevant moment.''
&
Evidence grounding and final decision
\\

\midrule

ReAct
&
``When reviewing the caption, carefully map each answer option to specific visual or audio events and their sequence.
\ldots identify any ambiguity \ldots formulate a single, highly
specific perceptual question \ldots''
&
Caption acquisition, query formulation, etc.
\\

\bottomrule
\end{tabular}
}
\end{table}

\vspace{-0.2cm}
\subsection{APO on the Agent-subagent Interface}
\label{ssec:results_react}
\vspace{-0.2cm}

We implemented the agent-subagent system as a ReAct-style pipeline
\cite{yao2022ReAct}, and applied GEPA. As shown in Table \ref{tab:react}, its
performance remains comparable to direct Gemini inference, while
GEPA does not improve held-out accuracy.

To understand the behavior of the agentic interface, we inspect
samples whose correctness changes between direct inference and ReAct.
Among direct-correct but ReAct-wrong cases, the dominant failures are
conflicting perception evidence (61.7\%) and incorrect reasoning over
otherwise useful evidence (27.3\%). Conversely, ReAct improvements
mainly come from retrieving additional useful evidence or richer
captions. Thus, active evidence acquisition in the agent system can recover missing information, but can also introduce extra failure points.

When applying GEPA on the agentic system, it succeeded in finding an optimized prompt with improved validation accuracy during prompt search, but the improvement does not transfer to test. This discrepancy suggests that the agentic interface is difficult to optimize reliably.
Table \ref{tab:optimized_prompt_examples} further shows that GEPA
acts on different scopes in the two interfaces. The optimized direct
prompt mainly constrains evidence grounding at the final decision,
whereas the optimized ReAct prompt changes a longer chain involving
caption interpretation, query formulation, evidence acquisition, and
integration. Accordingly, GEPA substantially changes the interaction
strategy, reducing the calls to query-responder by 35.8\% and total tool calls by 15.0\%, yet yielding no gains in held-out accuracy. There are few potential explanations. One is that final-answer correctness alone may provide a weak optimization signal for a multi-stage agentic interface. Or the perception and reasoning ability of the LLM components set limits on how the agentic system can perform.

\vspace{-0.2cm}
\section{Conclusions}
\label{sec:conclusions}
\vspace{-0.2cm}
We study whether externalize audio-visual reasoning of OmniLLM-based systems through explicit textual interfaces makes APO more effective than directly optimizing an OmniLLM instruction,which we found is not uniformly the case. APO substantially improves direct Gemini inference, reaching 79.62\% overall accuracy on Daily-Omni, despite it not producing much gain on Qwen3-Omni models. In contrast, the captioner-reasoner and agentic systems expose more interpretable optimization interfaces but obtain no reliable improvement from APO. Our error and trajectory analyses reveal interface-specific failure modes, including information loss in captioning, conflicting perception evidence, and downstream reasoning errors. Our results suggest that explicit textual interfaces make multimodal reasoning easier to inspect and steer, but not necessarily make APO more effective.


\section{Acknowledgments}
\label{sec:Acknowledgments}
This work was supported by the UKRI Centre for Doctoral Training (CDT) in Natural Language Processing, funded by the UKRI grant EP/S022481/1 and the University of Edinburgh. We would like to thank Dr. Yiwen Shao, Dr. Rohit Saxena, and Dr. Pasquale Minervini for useful discussion.

Generative AI tools were used to assist with language editing and grammar checking of the manuscript, as well as with code development and visualization. Any AI-generated content used in this work was thoroughly reviewed, edited, and verified by the authors. All scientific content, analyses, interpretations, and findings were developed and verified by the authors, who take full responsibility for the content of the manuscript.

\bibliographystyle{IEEEbib}
\bibliography{references}

\end{document}